\pdfoutput=1
\documentclass[11pt]{article}

\usepackage[final]{acl}

\usepackage{times}
\usepackage{latexsym}

\usepackage[T1]{fontenc}
\usepackage[utf8]{inputenc}

\usepackage{microtype}

\usepackage{inconsolata}

\usepackage{graphicx}

\usepackage{booktabs}

\usepackage{makecell}

\usepackage{float}

\usepackage{amsmath}

\title{Neural network embeddings recover value dimensions \linebreak from psychometric survey items on par with human data}

\author{
 \textbf{Max Pellert\textsuperscript{1}},
 \textbf{Clemens M. Lechner\textsuperscript{2}},
 \textbf{Indira Sen\textsuperscript{3}},
 \textbf{Markus Strohmaier\textsuperscript{3,4}}
\\
\\
 \textsuperscript{1}Barcelona Supercomputing Center
 \textsuperscript{2}Jam Technologies GmbH\\
 \textsuperscript{3}University of Mannheim\\
 \textsuperscript{4}GESIS - Leibniz Institute for the Social Sciences
\\
 \small{
   \textbf{Correspondence:} \href{mailto:max.pellert@bsc.es}{max.pellert@bsc.es}
 }
}

\begin{document}
\maketitle
\begin{abstract}
We demonstrate that embeddings derived from large language models, when processed with "Survey and Questionnaire Item Embeddings Differentials" (SQuID), can recover the structure of human values obtained from human rater judgments on the Revised Portrait Value Questionnaire (PVQ-RR). We compare multiple embedding models across a number of evaluation metrics including internal consistency, dimension correlations and multidimensional scaling configurations. Unlike previous approaches, SQuID addresses the challenge of obtaining negative correlations between dimensions without requiring domain-specific fine-tuning or training data re-annotation. Quantitative analysis reveals that our embedding-based approach explains 55\% of variance in dimension-dimension similarities compared to human data. Multidimensional scaling configurations show alignment with pooled human data from 49 different countries. Generalizability tests across three personality inventories (IPIP, BFI-2, HEXACO) demonstrate that SQuID consistently increases correlation ranges, suggesting applicability beyond value theory. These results show that semantic embeddings can effectively replicate psychometric structures previously established through extensive human surveys. The approach offers substantial advantages in cost, scalability and flexibility while maintaining comparable quality to traditional methods. Our findings have significant implications for psychometrics and social science research, providing a complementary methodology that could expand the scope of human behavior and experience represented in measurement tools.
\end{abstract}

\section{Introduction}

Psychometrics, the field of study in psychology that is crucially concerned with measurement of human attributes such as skills, personality or values, is undergoing a revolution with the integration of novel NLP technologies to complement and extend traditional questionnaire-based workflows. Proposed ideas range from the assessment of large language models via "AI Psychometrics" \cite{pellertAIPsychometricsAssessing2024} to "Machine Psychology" \cite{hagendorffMachinePsychologyInvestigating2023b} or, more conservatively, to use LLMs as assisting tools in human psychometrics to aid comparably cheap, in-silico questionnaire item creation \cite{gotzLetAlgorithmSpeak2023} or automated coding \cite{nilssonAutomaticImplicitMotive2025}. For the closely related task of synthetic data generation \cite{hommelTransformerBasedDeepNeural2022}, LLMs have been used to simulate segments of the population to match survey results \cite{argyleOutOneMany2023a} over time \cite{ahnertExtractingAffectAggregates2025} and to replicate experimental studies \cite{10.5555/3618408.3618425,dillionCanAILanguage2023}.

% Traditionally, the field of psychometrics is concerned with the careful creation and validation of the right set of items for correct measurement of the phenomenon of interest. 

The social and behavioral sciences crucially depend on texts that have been rated by human respondents. In psychology this is the traditional approach to analyze traits that are not directly observable such as value orientations \cite{schwartzMeasuringRefinedTheory2022}, personality  \cite{johnParadigmShiftIntegrative2008} and attitudes towards issues like gender beliefs \cite{kerrDevelopmentGenderRole1996a,hillDevelopmentValidationGenderism2005,schudsonGenderSexDiversity2022}, but the same holds also for fields like political science, sociology and human-computer interaction. Questionnaires, the instruments used to collect human rater judgments for measurement, are typically developed iteratively. First, researchers compile a large set of candidate \textit{items} based on theories of human experience and behavior. Items are short statements or questions believed to measure an underlying phenomenon, which  humans rate on verbal and/or numeric response scales, such as a five-point agree–disagree scale.  Through iterative procedures of elimination and refinement, researchers ultimately arrive at a final set of questionnaire items in which each item ideally measures only one underlying dimension of the phenomenon of interest (a property called "simple structure"). In this process, descriptive statistics and correlations among items obtained from samples of human raters are of paramount interest \cite{repkeValiditySurveyResearch2024}. Especially the patterns of inter-correlations are crucial in identifying underlying dimensions measured by the items. However, obtaining sufficiently large samples of human raters, and letting them rate a large number of candidate items, is often costly and plagued by issues such as respondent fatigue, response style, low literacy or lack of motivation. These problems can compromise data quality and thwart a proper assessment of the questionnaire's properties. We argue that addressing the challenge of obtaining fast, cheap and ideally unbiased information on the structure of questionnaires is a promising application area for NLP research.

Although the potential is large, critical evaluation has shown various limitations of existing attempts to marry NLP technologies with social scientific applications such as psychometric measurement \cite{lohn-etal-2024-machine-psychology,suhrChallengingValidityPersonality2024}. We review selected representative approaches by other researchers and discuss the problems they face. Then, we present our novel methodological advancement of "Survey and Questionnaire Item Embeddings Differentials" (SQuID) and demonstrate with the Revised Portrait Value Questionnaire (PVQ-RR) how it allows us to overcome issues that we identified.

\textbf{Research objective:} We want to demonstrate the power of neural embeddings in psychometric workflows with the example of the PVQ-RR. Furthermore, we want to argue for this approach to provide useful and interesting complements to human rater judgments.

\textbf{Approach:} We use LLMs to create semantic embeddings of the text of psychometric questionnaire items. We choose a widely used survey on value orientations that was carefully developed and extensively validated to compare our embeddings approach to traditional human rater judgments \cite{schwartzMeasuringRefinedTheory2022}.

\textbf{Results and contributions:} We show that, unlike some of the related works we review claim, semantic embeddings from a general-purpose model in combination with our methodological advancement can indeed recover very well the latent dimensions of a social scientific questionnaire. According to conventionally used quantitative evaluation metrics, we achieve a performance that is at least on par with methods relying on human rater judgments. Our robust findings demonstrate the feasibility and the potential of the "semantic embeddings as human rater judgments" approach and open a large domain of interesting research questions that can be tackled in powerful novel ways by adapting well-established classical psychometric analysis pipelines. While maintaining the same level of quality, our approach based on semantic embeddings is highly flexible, cheap and scales very well.

% With the example of Schwartz value questionnaire, we demonstrate that we can recover latent dimensions as well with human embeddings than with human data (cannonical, used for the theory development).

% or higher level?

\section{\label{sec:related}Related Work}

Researchers have proposed various ways of integrating LLMs in social scientific workflows. Typically, these efforts rely on repeatedly gathering stochastic model output in web interfaces of commercial, closed black-box APIs or on using older, smaller open models \cite{binzUsingCognitivePsychology2023,fischerWhatDoesChatGPT2023,serapio-garciaPersonalityTraitsLarge2025}. In this section, we review the most relevant recent literature that we found to be so far in the domain of personality questionnaires.

\textbf{First uses of embeddings to test the structure of psychometric inventories}

A number of recent studies have proposed using embeddings of questionnaire items from psychometric inventories—or various aggregates of items. These embedding vectors can then be used to create similarity matrices consisting of pairwise similarities between all items or item aggregates. These, in turn, can be analyzed with factor-analytic or similar dimension-reduction methods in much the same way that researchers have analyzed correlation matrices resulting from human subjects' responses to psychometric inventories. 
On a fundamental level that also ensures full reproducibility, \citet{cutlerDeepLexicalHypothesis2023a} successfully replicated a classical human subject study on personality traits  using an embedding approach. The authors created templates: "Those close to me say I have a [MASK][MASK] and [TERM] personality." [TERM] is inserted from a list of personality descriptors containing entries such as "condescending", the constructed statement is passed through the model and the token embeddings of the two [MASK] tokens are extracted and averaged. The results of principal component analysis on that averaged embedding were reasonably similar to results of a classical study on human ratings of personality descriptors \cite{saucierEvidenceBigFive1996} with 3 out of 5 factors adequately well recovered. The authors used a less systematic and more combinatorial approach of running the same experiment with a large number of different models, all versions of BERT, to find the best-performing one according to factor congruence scores with the original human rater judgments. Interestingly, a model finetuned for Natural Language Inference (NLI) tasks, the head of which was in fact discarded for this setup (as only the two masked token embeddings are extracted), performed best. Cutler and Condon did not investigate why this model worked best. Moreover, their template-based approach is largely limited to personality inventories, for which this structure is adequate. While their replication study did not provide an explicit acknowledgment of the generality of the approach, it presented  pioneering evidence of the power of neural embeddings in emulating human rater judgments.

\textbf{Embedding items of different questionnaires to compare them}

Researchers have also taken a more macro-perspective and worked with embeddings of data sets that collect a number of different personality questionnaires. \citet{wulffSemanticEmbeddingsReveal2025a} finetune a BERT based model on 200,000 pairs of items and their empirical correlations to create a domain-specific embedding model. Laudably, they share their model publicly on the Hugging Face Model Hub \footnote{\href{https://huggingface.co/dwulff/mpnet-personality}{https://huggingface.co/dwulff/mpnet-personality}} and we include it in our analyses. They use their model to embed a large number of different personality items, scales and construct names. These embeddings allow them to predict empirical similarities between items reasonably well, to detect overlaps between scales and to suggest more parsimonious taxonomies. Relying on a very similar base model, \citet{hommelLanguageModelsAccurately2025} used finetuning to create an embedding model that allows them to establish inter-item correlations, scale reliability and inter-scale correlations to some degree. Unfortunately, their model is not openly shared and therefore cannot be used by us or by other researchers.

\textbf{First proposals of embeddings as part of the psychometric toolbox}

These approaches are generalized by \citet{feracoSEMbeddingsHowEvaluate2025} by building on the before-mentioned "mpnet-personality" model to create a pipeline module that allows researchers to run in-silico validity checks on item candidates to identify problems in a cheap and easy way before any empirical data from  human raters is collected. Similarly, "pseudo-factor analysis" was recently proposed as a pre-testing approach that allows evaluating the semantic structure of personality inventories (and potentially other psychological constructs)  \cite{guenoleEnhancingScaleDevelopment2025}. Like the approach by Feraco and colleagues, pseudo-factor analyis is based on embedding items or parcels of items and computing similarity matrices between embedding vectors, which can then be factor-analysed in the same way as inter-item correlations of personality ratings from human respondents are traditionally factor analyzed. Guenole and colleagues illustrated their approach with two widely used personality inventories, reporting that pseudo-factor analyses was able to replicate empirical findings on the structure of these inventories obtained on human subject data reasonably well, at least when using (the rather permissive) oblique rotation. Despite these recent advances, it remains unclear which of the many different models and strategies for creating and combining embeddings on different items or item parcels works best, let alone why.

\textbf{Missing negative correlations in embeddings of psychometric inventories}

In personality inventories and other psychometric inventories, it is common practice to use both forward-keyed (e.g., "I am meticulous.") and reverse-keyed items (e.g., "I am rather careless.") to measure the same personality construct.  Such item pairs would be expected to correlate negatively with each other yet load on the same factor (e.g. Conscientiousness). Likewise, constructs and their subdimensions, such as the Big Five dimensions Neuroticism and Extraversion, often correlate negatively with each other. Such negative correlations are important properties of both the semantic space of psychological constructs and the empirical data resulting from human responses to personality inventories.
However, \citet[p.7]{hommelLanguageModelsAccurately2025} reported an important problem: "negative coefficients are rarely produced when comparing vector representations of sentences". This is evidenced by the scatterplot of empirical and their predicted inter-item similarities using embeddings which are cutting of at 0. This problem—shared by all of the abovementioned recent embeddings-based approaches—stems from the high-dimensional embeddings of item text containing much information on general abstract linguistic features of natural language use that are similar even across otherwise semantically distinctly dissimilar item texts. The inability to reproduce negative correlations between items  constitutes a major impediment for reproducing patterns of correlations and factor loadings obtained in human subjects through embeddings-based approaches. 
Hommel and Arslan developed a workaround that involves re-annotating an NLI corpus \cite{N18-1101} by assigning negative direction to predicted similarity when the example pair is labeled as contradictory in the corpus. They then fine-tune their model on this adapted data. This workaround seems to have the effect of increasing performance for their use case because the model better learns to handle negation and contradictions between item texts. Different to that, Wulff and Mata decide to fully drop the sign of the correlation and just consider "unsigned (absolute) correlation as a measure of item relatedness" \citeyearpar[p.1 of SI]{wulffSemanticEmbeddingsReveal2025a}. This approach can work for the authors because all questionnaires that are analyzed by them are in the personality domain and, therefore, exhibit a substantial similarity between each other to start from. The \textit{degree} of similarity still provides important information. Also,  Wulff and Mata pre-process items by removing the starting pronoun "I", a form of stemming which can be expected to remove one source of confounding shared abstract linguistic features and thereby reduce similarities.

Different to these prior attempts, we aim for a general methodology that can readily be applied post-hoc to embeddings of diverse contexts also different from personality derived from \textbf{any} model without re-annotating training corpora and finetuning. As described in Section \ref{subsec:ses}, we develop "Survey and Questionnaire Item Embeddings Differentials" (SQuID) as a general method that fulfills these constraints and allows obtaining negative correlations.

\section{Background}

\subsection{The theory of values}

All the works discussed in Section \ref{sec:related} have so far been in the domain of questionnaires for personality traits and attendant inventories (e.g., Big Five and HEXACO), which are a natural starting point because of the long research tradition of the 'lexical approach' investigating the semantic structure of personality-descriptive adjectives and statements \cite{guenoleEnhancingScaleDevelopment2025}. However, for embedding-based approaches to be a truly valuable addition to the psychometric toolkit, their viability needs to be demonstrated on additional constructs beyond personality traits. In our study, we therefore aim to go beyond personality inventories  and choose a widely used and well-established survey for the important domain of values. According to Schwartz’s theory of basic human values, values are trans-situational goals that serve as guiding principles in people's lives, expressing distinct motivational concerns. The "Revised Portrait Value Questionnaire" (PVQ-RR) \cite{schwartzRefiningTheoryBasic2012} contains 57 items such as "It is important to him/her to form his/her views independently.", "It is important to him/her that his/her country is secure and stable.", "It is important to him/her to have a good time.", which can be aggregated to 19 fine-grained value dimensions such as "Self-direction Thought", "Security-Societal" and "Hedonism" (3 items per dimension). Human respondents rate how similar they view themselves to the person described in the item on a 6-point scale ranging from "not like me at all" to "very much like me". The survey scoring rules that specify the dimension that each item belongs to and can be aggregated to through averaging have been developed in a theory-guided way that has been cross-culturally validated and refined since the 1980s with human rater judgments from 49 countries \cite{schwartzMeasuringRefinedTheory2022}. 

\subsection{Fundamental assumptions when adapting psychometric workflows to LLMs}

To empirically test their theory, Schwartz et al. collect a large number human rater judgments on a pre-specified response scale, for each of the 57 item of the PVQ-RR: $\mathbf{y}_i = (r_{i1}, r_{i2}, \ldots, r_{in})$ where each $r_{ij}$  represents the rating given by human $j$ for item $i$.
In principle, this data is in the same format as the item embeddings that we will extract from the models: $\mathbf{y}_i = (e_{i1}, e_{i2}, \ldots, e_{in})$ where each $e_{ij}$  represents the element $j$ of the embedding vector for item $i$. Both types of data can be used in the same way in downstream psychometric analysis pipelines (for a schematic overview see Figure \ref{fig:workflow}).

\subsection{Multidimensional Scaling and the Centrality of Negative Correlations }

Importantly, negative correlations between items play a very important role in Schwartz's theory of the content and structure of human values. Schwartz's values are structured in a circumplex model based on the compatibility and conflict between underlying motivations, whereby values that motivationally oppose each other are placed on opposite positions of the circle and correlate negatively with each other (e.g. Conformity vs. Self-direction), whereas closely related values that are motivationally compatible with each other are positively correlated and located close to each other on the same side of the circle (e.g. Achievement and Power). The individual values can be further aggregated to four higher-order dimensions corresponding to four segments of the circle that emerge with great regularity in data from human respondents \cite{lechnerMeasuringFourHigherOrder2024}.

This circular structure is uncovered using a dimension-reduction technique that is different from principal component analysis and factor analysis as typically used for personality inventories: Multi-dimensional scaling (MDS) \cite{jungMultidimensionalScaling2015,groenenMultidimensionalScalingII2015}. Schwartz's theory and attendant inventories were historically developed in close relationship with this technique. MDS maps high-dimensional objects such as vectors of human rater judgments or semantic embeddings to points in a lower-dimensional, typically 2D, space. The technique works in such a way that points corresponding to similar objects are located close together, while dissimilar objects are located far apart, thereby preserving distances. The MDS configurations that are computed from human rater judgments confirm those relationships between values as postulated by the theory. We can therefore empirically validate the theoretically assumed circular placement of values using MDS with human rater judgments \cite{schwartzMeasuringRefinedTheory2022}, while, until now, it was an open question if the same is possible with an embedding approach. The paramount importance of negative correlations between items renders Schwartz's value theory and the PVQ-RR an ideal (though challenging) case to refine embeddings-based approaches such that they reproduce negative correlations between items foundational to value theory.

\section{Experimental Setup}

\subsection{\label{subsec:embedd_models}Embedding Models}

Rapid development makes systematic model reviews quickly outdated. We therefore strategically selected models representative of specific model classes and types. This way, we can provide some guidance on which models typically to look for in this application area also in the future.

First, we choose the currently best performing open model for semantic textual similarity (STS) tasks in "MTEB: Massive Text Embedding Benchmark" \cite{muennighoff_mteb_2023}, \href{https://huggingface.co/Linq-AI-Research/Linq-Embed-Mistral}{Linq-Embed-Mistral} (7.11B, prompt,  \citealt{choiLinqEmbedMistralTechnicalReport2024}). In addition, we choose the currently best performing closed API model in the benchmark, \href{https://developers.googleblog.com/en/gemini-embedding-text-model-now-available-gemini-api/}{gemini-embedding-exp-03-07} (parameter count not disclosed, no prompt). Furthermore, we include two smaller models created with innovative training and data strategies that allow them to perform well for their size, \href{https://huggingface.co/jinaai/jina-embeddings-v3}{jina-embeddings-v3} (572M, no prompt, \citealt{sturuaJinaembeddingsv3MultilingualEmbeddings2024a}) and \href{https://huggingface.co/HIT-TMG/KaLM-embedding-multilingual-mini-instruct-v1.5}{KaLM-embedding-multilingual-mini-instruct-v1.5} (494M, prompt, \citealt{huKaLMEmbeddingSuperiorTraining2025}). Finally, we include the model \href{https://huggingface.co/dwulff/mpnet-personality}{mpnet-personality} (109M, no prompt) that, as discussed before, was created through domain specific finetuning. Our approach does not need extensive prompt engineering, any sophisticated token probability extraction strategies or accounting for stochasticity of outputs. In a growing social scientific literature of using LLMs for empirical investigations, instead of interacting with models on the textual output level we show the advantages of working on the fundamental level of semantic embeddings. We use the exact wording from the PVQ-RR questionnaire specifications (see Subsection \ref{subsec:prompt}) for those models that allow including a prompt\footnote{We run all models on CPU to embed all 57 items in less than 20 minutes on a standard ThinkPad T480 from 2018.}. The PVQ-RR is available in gendered version using different pronouns. We account for this feature by creating an average of the female and male item embeddings in order to reduce possible gender biases. 

As we go through in detail in Subsection \ref{subsec:validity}, we find evidence that the MTEB benchmark is actually a good guide for selecting models for this task. There is reason to expect that future models that score even better on the benchmark will likely still outperform our already strong results. 

\subsection{\label{subsec:ses}Survey and Questionnaire Item Embeddings Differentials (SQuID)}

Working with raw embeddings as directly retrieved from the models, we faced the same problem of missing negative correlations as the other researchers whose work we discussed in Section \ref{sec:related} (see the similarity matrix on the left of Figure \ref{fig:scalesubtraction}). As we will use MDS for dimensionality reduction which crucially builds on preserving distances, it was especially important for us to find a way to make the signal of negative correlations stronger.

As discussed before, the root of the problem lies in the high-dimensionality of item embeddings that encode many common linguistic features of natural language use which artificially boost similarity between items. To create a vector that captures this unneeded information, we take an average embedding over all items (a "questionnaire embedding") and subtract it from each of the item embeddings: $\mathbf{y}_i-\overline{\mathbf{y}}$ for item $i$, where $\overline{\mathbf{y}}$ is an average over all 57 items of the PVQ-RR: $\overline{\mathbf{y}} = \frac{1}{57}\sum_{i=1}^{57}\mathbf{y}_i$. This straightforward method is inspired by the early examples of vector arithmetic demonstrated with Word2Vec \cite{mikolovEfficientEstimationWord2013} and the more recent idea of community embeddings \cite{wallerQuantifyingSocialOrganization2021}. As can be seen in the right hand part of Figure \ref{fig:scalesubtraction}, this approach is simple yet very effective in making the signal of negative correlations appear without any domain-specific finetuning or other adaption procedures. SQuID works with any model and embeddings of arbitrary dimensions, offering a high degree of flexibility. We believe that a substantial part of the under-performance of embedding approaches compared to human data in other studies, is due to methodological reasons and can be overcome with SQuID.

% Cannot replace (and there are strong reasons why they should never) yet but r

\begin{table*}[!t]
\fontsize{12.0pt}{14.4pt}\selectfont
\begin{tabular*}{\linewidth}{@{\extracolsep{\fill}}l|rrrrrrrrrr}
\toprule
 & AC & BEC & BED & COI & COR & FAC & HE & HUM & POD & POR \\ 
\midrule\addlinespace[2.5pt]
gemini & 0.45 & 0.74 & 0.71 & 0.80 & 0.74 & 0.47 & 0.63 & 0.34 & 0.74 & 0.74 \\ 
jina & 0.60 & 0.77 & {\bfseries 0.76} & 0.73 & 0.77 & 0.51 & 0.69 & 0.22 & 0.58 & 0.83 \\ 
kalm & 0.61 & 0.72 & 0.70 & 0.76 & 0.79 & 0.34 & 0.62 & 0.23 & 0.64 & 0.81 \\ 
linq & {\bfseries 0.72} & {\bfseries 0.89} & 0.68 & {\bfseries 0.83} & {\bfseries 0.90} & 0.53 & 0.72 & 0.46 & {\bfseries 0.82} & 0.87 \\ 
mpnet & 0.37 & 0.74 & 0.52 & 0.68 & 0.75 & 0.18 & 0.62 & -0.05 & 0.63 & {\bfseries 0.88} \\ 
$\mathcal{N}(0,\,1)$ & -0.06 & -0.06 & -0.06 & -0.04 & -0.06 & -0.04 & -0.04 & -0.04 & -0.04 & -0.06 \\ 
human & 0.60 & 0.74 & 0.69 & 0.75 & 0.81 & {\bfseries 0.68} & {\bfseries 0.73} & {\bfseries 0.47} & 0.73 & 0.76 \\ 
\bottomrule
\end{tabular*}

\vspace{0.5em}

\fontsize{12.0pt}{14.4pt}\selectfont
\begin{tabular*}{\linewidth}{@{\extracolsep{\fill}}l|rrrrrrrrrr}
\toprule
 & SDA & SDT & SEP & SES & ST & TR & UNC & UNN & UNT & \multicolumn{1}{c}{$\mu$} \\ 
\midrule\addlinespace[2.5pt]
gemini & 0.78 & 0.57 & 0.54 & 0.67 & 0.65 & 0.70 & 0.53 & 0.66 & 0.72 & 0.64 \\ 
jina & 0.78 & 0.66 & 0.49 & 0.81 & 0.73 & 0.76 & 0.49 & 0.79 & 0.73 & 0.67 \\
kalm & 0.79 & {\bfseries 0.68} & 0.62 & 0.68 & 0.71 & 0.82 & 0.63 & 0.85 & 0.71 & 0.67 \\ 
linq & {\bfseries 0.82} & {\bfseries 0.68} & {\bfseries 0.66} & {\bfseries 0.91} & {\bfseries 0.78} & {\bfseries 0.83} & {\bfseries 0.77} & {\bfseries 0.92} & {\bfseries 0.85} & {\bfseries 0.77} \\ 
mpnet & 0.55 & 0.36 & 0.50 & 0.58 & 0.72 & 0.81 & 0.49 & 0.86 & 0.63 & 0.57 \\ 
$\mathcal{N}(0,\,1)$ & -0.06 & -0.04 & -0.06 & -0.04 & -0.06 & -0.06 & -0.04 & -0.04 & -0.06 & -0.05 \\ 
human & 0.66 & 0.65 & 0.58 & 0.77 & 0.67 & 0.80 & 0.73 & 0.83 & 0.71 & 0.70 \\ 
\bottomrule
\end{tabular*}
\caption{\label{tab:cronbach}\textbf{Internal consistency as measured through Cronbach's alpha for each of the 19 value dimensions and the average over dimensions.} Rowname abbreviations refer to the models covered in Subsection \ref{subsec:embedd_models} (all embeddings were SQuID treated). We report a random baseline of embeddings created from a normal distribution with mean zero and unit variance (averages over 10 000 samples). Human data is retrieved from Table S2 in SI of \citet{schwartzMeasuringRefinedTheory2022}. Linq-Embed-Mistral is generally the best performing embedding model with an average value for Cronbach's alpha that is above even the human benchmark. Values close to 0.8 or above for most value dimensions point to adequate internal consistency (with notable exceptions such as HUM/humility, which is also low in the human data). For fully spelled out dimension names in the column header see Figure \ref{fig:scalesubtraction}.}
\end{table*}

% embeddings created from \href{https://developers.googleblog.com/en/gemini-embedding-text-model-now-available-gemini-api/}{gemini-embedding-exp-03-07}, \href{https://huggingface.co/jinaai/jina-embeddings-v3}{jina-embeddings-v3} \cite{sturuaJinaembeddingsv3MultilingualEmbeddings2024a}, \href{https://huggingface.co/HIT-TMG/KaLM-embedding-multilingual-mini-instruct-v1.5}{KaLM-embedding-multilingual-mini-instruct-v1.5} \cite{huKaLMEmbeddingSuperiorTraining2025}, \href{https://huggingface.co/Linq-AI-Research/Linq-Embed-Mistral}{Linq-Embed-Mistral} \cite{choiLinqEmbedMistralTechnicalReport2024} and \href{https://huggingface.co/dwulff/mpnet-personality}{mpnet-personality}

\section{Results}

Unless otherwise noted, the results presented here are derived from SQuID treated embeddings from the best-performing model "\href{https://huggingface.co/Linq-AI-Research/Linq-Embed-Mistral}{Linq-Embed-Mistral}". For performance of all models compared to the human rater judgment data as measured with factor congruence scores, see Table \ref{tab:congruence_models}.

\subsection{\label{subsec:validity} Validity checks}

% Traditionally, semantic embeddings are evaluated on semantic similarity benchmarks \cite{muennighoff_mteb_2023} consisting of curated datasets of found data such as \href{https://huggingface.co/datasets/sentence-transformers/reddit-title-body}{labels of subReddits with (title, body) pairs}. Our twofold evaluation procedure is different from that and adds to the NLP literature on validating semantic similarity tasks.

\subsubsection{Internal consistency}

Following \citet{schwartzMeasuringRefinedTheory2022}, we employ Cronbach's alpha to assess the internal consistency of latent dimensions. Through functional equivalence in this context (for details see Subsection \ref{subsec:cronbach}), we can directly compare embeddings of all our candidate models with human data\footnote{We retrieve the human rater judgment data from Table S2 in SI of \citet{schwartzMeasuringRefinedTheory2022} which reports averages over 49 countries.}. Cronbach's alpha compares the amount of shared variance among the items of a dimension to the amount of overall variance. In previous research, performance baselines of random embeddings have been shown to be higher than intuitively expected in some scenarios \cite{wietingNoTrainingRequired2019}. To assess which baseline levels we reach, we produce a vector with 4096 normally distributed values ($M_{ij} \sim \mathcal{N}(0,\,1)$) for each item. We repeat the procedure 10 000 times and report average values of Cronbach's alpha for each dimension in these random embeddings. In addition to reporting all 19 values for each model, human data and random embeddings, we also provide an overall average over all dimensions. In Table \ref{tab:cronbach}, we find that Linq-Embed-Mistral is the best performing model. It beats the other approaches 14 out of 19 times and has the highest average Cronbach's alpha over all dimensions (including being slightly higher than the one measured from the human data).

\subsubsection{Dimension-dimension correlations of human rater judgments and embeddings}

\begin{figure}[htpb!]
  \includegraphics[width=\columnwidth]{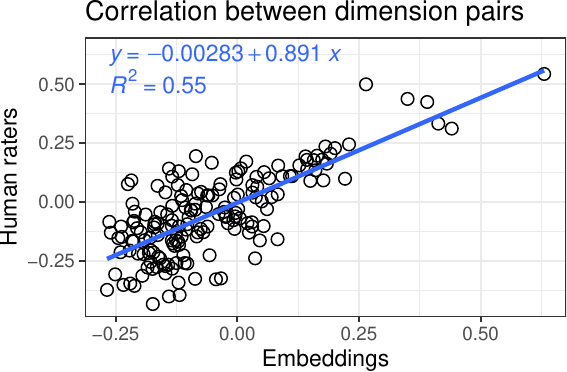}
  \caption{\label{fig:scatterplot} \textbf{Relationship between item similarity (Pearson correlation) derived from embeddings (SQuID treated) and human data.} We explain more than half of the variance with a simple linear model.}
  \vspace{-10pt}
\end{figure}

We create a dimension-dimension similarity matrix (using Pearson correlation) for both the embeddings (shown in Subsection \ref{subsec:corrplot_dimensions}) and the human data. The 171 pairs of both types of data positively correlate substantially: $0.74$ [$0.66,0.80$]. Figure \ref{fig:scatterplot} shows the scatter plot and parameters of a simple linear model that explains $55\%$ of the variance observed.

\subsection{MDS configuration from embeddings}

\begin{figure}[!htpb]
\centering
  \includegraphics[width=0.65\columnwidth]{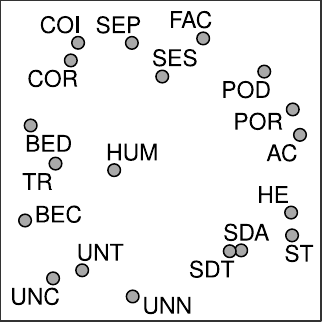}
  \caption{\label{fig:mdsconfiguration} \textbf{Unrotated MDS configuration derived from SQuID treated embeddings.} As value dimensions are placed according to their distances to each other, a circular structure emerges. For fully spelled out dimension names see Figure \ref{fig:scalesubtraction}.}
  \vspace{-10pt}
\end{figure}

We turn the dimension-dimension correlation matrix of Subsection \ref{subsec:corrplot_dimensions} into a dissimilarity matrix. We use multidimensional scaling\footnote{\textbf{mds} with 'type="ordinal"' from the R package smacof \cite{mairMultidimensionalScalingUsing2022}, for details see \href{https://github.com/maxpel/embeddings_values/blob/main/05_multidimensional_scaling.R}{code file}.} to reduce the high-dimensional dissimilarity matrix to a two dimensional space that preserves distances between dimensions. Figure \ref{fig:mdsconfiguration} shows the resulting circular structure. 

\subsection{Procrustes matrix transformation to compare human data and embeddings}

Next, we want to compare the embeddings MDS configuration to the empirical configuration of \citet{schwartzMeasuringRefinedTheory2022}\footnote{We compute the human MDS configuration from the pooled correlation matrix over all 49 countries shared by the authors as Table S10 of SI.} derived from human rater judgments. As MDS configurations are not unique, arbitrary matrix transformations such as translation, rotation and reflection are admissible because these operations do not change pairwise distances. We use Procrustes matrix transformation\footnote{\textbf{procrustes} from the R package \href{https://cran.r-project.org/web/packages/vegan/index.html}{vegan}} with the human configuration as target and the embeddings configuration as testee. This way, we transform the embeddings configuration to achieve maximum similarity with the human rater judgments configuration. To quantitatively assess the similarity of the two MDS configurations after Procrustes transformation of the embeddings MDS configuration matrix, we compute cosine similarities of the vectors of the two first dimensions (X coordinates in the configuration plot) as well as the two second dimensions (Y coordinates). These congruence coefficients are $0.88$ for the first and $0.82$ for the second dimension. To further assess similarities and differences between individual dimensions, Figure \ref{fig:comparisonmds} allows for visual assessment of the two configurations.

\vspace{-1pt}

\subsection{Generalizability to other questionnaires}

To assess whether SQuID generalizes beyond value theory, we apply the method to three widely-used personality questionnaires: the International Personality Item Pool (IPIP) \cite{goldbergDevelopmentMarkersBigFive1992}, the Big Five Inventory-2 (BFI-2) \cite{sotoNextBigFive2017}, and the HEXACO Personality Inventory \cite{leePsychometricPropertiesHEXACO1002018}. These personality measures differ from the PVQ-RR in both theoretical structure and item content, providing a rigorous test of SQuID's domain generality.
Similar to value theory, a challenge in personality measurement is capturing negative correlations between items with opposing content. The range of inter-item correlations thus serves as a critical metric: wider ranges indicate that embeddings successfully capture both positive associations between similar items and negative relationships between opposing items.
Table \ref{tab:model_comparison} shows that SQuID substantially increases correlation ranges across all models for the IPIP questionnaire. Raw embeddings exhibit limited ranges (0.14 to 0.82 depending on model), but SQuID treatment expands these dramatically. The average relative gain in range is 261.5\%, with individual models showing improvements from 21.20\% (mpnet, which was already optimized for personality content through domain-specific finetuning) to 724.39\% (kalm). Notably, even models that initially showed reasonable raw ranges (e.g., mpnet at 0.82) benefit from SQuID.
We observe similar patterns for BFI-2 (average relative gain over all models: 207.2\%) and HEXACO (147.9\%). These consistent improvements across multiple personality inventories suggest that SQuID can be applied broadly to psychometric measurement.

\begin{table}[!htbp]
\fontsize{12.0pt}{14.4pt}\selectfont
\begin{tabular*}{\linewidth}{@{\extracolsep{\fill}}l|rrrr}
\toprule
 & \makecell{Raw\\Range} & \makecell{SQuID\\Range} & \makecell{Abs.\\Gain} & \makecell{Rel.\\Gain (\%)} \\
\midrule\addlinespace[2.5pt]
gemini & 0.35 & 0.92 & 0.57 & 161.88 \\
jina & 0.66 & 1.30 & 0.64 & 95.97 \\
kalm & 0.14 & 1.19 & 1.04 & 724.39 \\
linq & 0.23 & 0.92 & 0.69 & 304.13 \\
mpnet & 0.82 & 1.00 & 0.17 & 21.20 \\
\bottomrule
\end{tabular*}
\caption{\label{tab:model_comparison}\textbf{Comparison of raw vs SQuID embeddings across all models for the IPIP personality questionnaire.} Raw Range and SQuID Range show correlation ranges. Absolute Gain shows the increase in range, while Relative Gain shows the percentage improvement.}
\end{table}

% We conduct a successful test of the generalizability of SQuID by using it on the item embeddings of other types of questionnaires from different domains. SQuID increases the range of inter-item correlations for all models on all additional questionnaires that we tested, on some substantially so. Table \ref{tab:model_comparison} shows the ranges for embeddings from all the models under consideration as-is, after employing SQuID and the relative gain of range for the IPIP personality questionnaire\cite{goldbergDevelopmentMarkersBigFive1992}. The average relative gain of range for IPIP is 261.5\%. Furthermore, we tested SQuID on the BFI-2\cite{sotoNextBigFive2017} and HEXACO\cite{leePsychometricPropertiesHEXACO1002018} questionnaires. We observe an average relative gain of range of 207.2\% and 147.9\% over all models.

\section{Discussion}

Our validation checks uncover that value dimensions contained in neural embeddings can display internal consistency on par with human data (or even above). As a comparison baseline, we find no meaningful values for randomly initiated embeddings. Correlation of dimension-dimension similarities for human data and embeddings show substantial agreement between the two types of data. We find no bias in regressing one on the other, a strong positive relationship and a moderately high coefficient of determination for a simple linear model.

In the embeddings configuration of Figure \ref{fig:mdsconfiguration}, we can see the circular structure emerging that is postulated by \citet{schwartzMeasuringRefinedTheory2022}. Values that are facets of the same phenomenon (like "Universalism-Concern", "Universalism-Nature" and "Universalism-Tolerance") are placed close together. Values that are seen as distant to Universalism (such as "Power Dominance" and "Power Resource") are placed far apart on the circular structure, roughly on the opposite side. The same can be observed with the facets of "Self-Direction" in comparison to the facets of "Conformity". The results allow for a visual interpretation that follows quite closely along the lines of the canonical theory of value orientations that powers the PVQ-RR.

\begin{figure}[htpb!]
\centering
  \includegraphics[width=0.83\columnwidth]{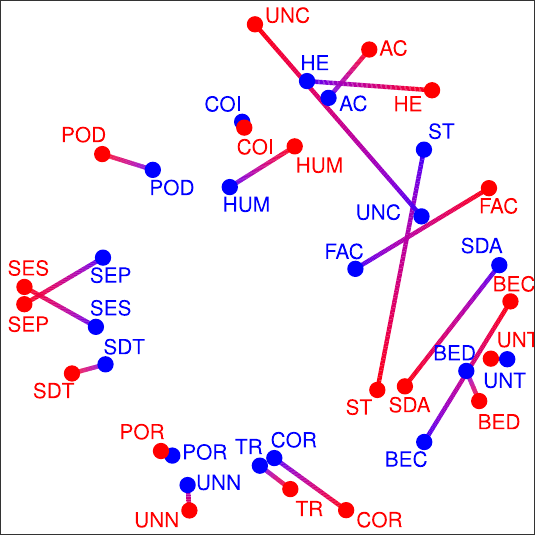}
  \caption{\label{fig:comparisonmds} \textbf{Comparison of MDS configurations from embeddings and from human data.} The configurations are shown after Procrustes similarity transformation. Human data in \textcolor{red}{red}, embeddings in \textcolor{blue}{blue}. We find few long and few strongly overlapping lines evidencing general similarity of the two configurations derived from two distinct types of data. For fully spelled out dimension names see Figure \ref{fig:scalesubtraction}.}
  \vspace{-10pt}
\end{figure}

Comparing the two MDS configurations after Procrustes transformation shows fair similarity according to conventional interpretations of congruence coefficients (for details see Section \ref{subsec:congruence}). That we find the two configurations to be fairly similar on this formal criterion is quite surprising given that they derive from two very distinct data sources. Inspection of individual dimensions in Figure \ref{fig:comparisonmds} reveals strong overlaps (e.g., Security facets, Conformity-Interpersonal, Power Resources) and some differences (e.g., Stimulation is placed closer to Hedonism in the embeddings configuration).

% Inspection of individual value dimensions in Figure \ref{fig:comparisonmds} reveals strong overlaps of the two solution with the facets of Security (Security Personal and Security Societal) and basically identical placement of some dimensions such as Conformity-Interpersonal, Power Resources and Universalism-Tolerance. While other are placed closely like Self-direction Thought and Power Dominance, dimensions can also be distinctly different between the two configurations, for example  Stimulation (ST) that is placed much closer to Hedonism (HE) in the embeddings configuration.

% Using SQuiD also made the the signal of negative correlations stronger for three questionnaires (IPIP, BFI-2, HEXACO) in the different domain of personality. These extension results demonstrate that SQuID generalizes. 

\section{Conclusion}

We introduced our methodological advancement of SQuID and show with the example of the PVQ-RR that it allows an embedding approach to reach performance on par with human data in extracting latent value dimensions according to several evaluation metrics. We envision many new directions of research to emerge from psychometric approaches to AI. As one example, our approach allows for a systematic and rigorous investigation of potential cultural biases contained in LLMs: \citet{schwartzMeasuringRefinedTheory2022} confirmed their theories across 49 countries using translated questionnaires. We see an exciting opportunity to use our approach and MDS configurations on the level of those individual countries to systematically assess closeness or distance of an LLM to certain cultural value orientations quantitatively by computing congruence coefficients and other error metrics. Working on the embedding level, as we propose, limits sources of potential methodological biases like token extraction strategies that are necessary when working with generated text instead.

The fundamental difference between human rater judgments and neural embeddings as data sources raises important questions about what each captures. Human ratings reflect conscious, individual self-assessment assumed to be influenced by the traits to be measured. In contrast, embeddings represent semantic relationships derived from statistical regularities in language use learned from vast text corpora by neural models. That these two data sources converge to produce similar psychometric structures (R² of 0.55, fair congruence coefficients) suggests that they nevertheless tap into shared underlying semantic spaces. The 45\% of variance not explained by our linear model could stem from genuine differences in what the two data types measure, psychometric noise in human responses (e.g. response styles, measurement error) or systematic biases in language model training data. Future research should decompose this unexplained variance to better understand the complementary information each data source provides.

While our results demonstrate SQuID's effectiveness for value orientations (PVQ-RR) and initial tests show promise for personality inventories (IPIP, BFI-2, HEXACO), systematic evaluation across additional psychometric domains remains an important direction for future research. We suggest several priority areas:
First, researchers should test SQuID on questionnaires with different structural properties. The PVQ-RR exhibits a circumplex structure suited to multidimensional scaling, while personality inventories typically employ factor-analytic approaches. Testing on hierarchical models, formative constructs or network-based psychometric structures would clarify boundary conditions.
Second, questionnaires employing reverse-keyed items require methodological refinement. As noted in our limitations, directly "reversing" embeddings is conceptually unclear. One possible approach could involve computing separate scale embeddings for regular and reverse-keyed items, then subtracting separately from the two types of items. This extension would at least better handle the intentional polarity inversions common in personality and attitude measures.
Third, extend testing to attitudes, motivational orientations and clinical inventories, which may require methodological adaptations.
Finally, multilingual testing would establish whether SQuID's benefits generalize across languages.

We view our methodological proposal as complementary to both traditional workflows as well as to novel approaches in the (social) sciences. By adopting embeddings approaches, we don't aim to replace human participants. Also well into the future, researchers will have good reasons to study humans, at the bare minimum to provide necessary reference points for comparison as we demonstrated in our analyses. But different to ideas discussed before, such as "pseudo-factor analysis" \cite{guenoleEnhancingScaleDevelopment2025}, we want to go beyond merely creating another step in the typical validation pipeline in psychometrics. We see potential beyond that: With these novel technologies we can potentially investigate the full and diverse spectrum of expression in all natural languages. This creates novel opportunities to expand the scope of human behavior and experience represented in measurement tools.

 % opens new doors. Performance .

% In this work, we are providing in-depth analysis of one questionnaire. Representative and methodologically eseciall interesting with multidimensional scaling. Not focusing across scales but future research should do, to use our methodological improvements to improve existing workflows.

% Value orientations specifically: opportunity to systematically establish value orientations without the need for extensive prompt engineering or sophisticated token probablity extraction techniques or just collecting textual outputs (with all thr arising issues).

% Lexical Hypothesis on steroids.

% Since December 2023, a "Limitations" section has been required for all papers submitted to ACL Rolling Review (ARR). This section should be placed at the end of the paper, before the references. The "Limitations" section (along with, optionally, a section for ethical considerations) may be up to one page and will not count toward the final page limit. Note that these files may be used by venues that do not rely on ARR so it is recommended to verify the requirement of a "Limitations" section and other criteria with the venue in question.

\section*{Limitations}

Future research will have to uncover more of the peculiarities of that novel kind of data in the form of semantic embeddings for (social) scientific applications. More experience with the data will in turn help the methods (and the theories). We see "neural embeddings as human rater judgments" as a good working hypothesis and don't want to imply that both data are exactly the same and capture human behavior and experience in exactly the same way. As with other synthetic data, novel treatments such as specific transformation (in addition to general methodologies like SQuID) and other processing steps may be necessary depending on the application context.

Questions of measurement validity arise: Do findings from the human domain hold generally? In this emerging field, we don't yet have established answers to these questions. We can face unexpected problems, as for example with questionnaire scoring rules that do prescribe to reverse certain items for data quality reasons. With human rater judgments this can be done in a straightforward way (by subtracting the rating from the maximum score possible as a response), but how do we "reverse" semantic embeddings in a meaningful way? Novel methodological questions like these appear often and have to be addressed in future research. With our work, we want to make a strong argument that it is worth attempting to provide answers.

The PVQ-RR is openly available on the internet and it could have been part of the training corpus of the models that we use. The models' direct memorization of its items and the underlying theory could potentially influence our results and limit the scope of our methodology when applied to different, less easily accessible contexts. We don't think that this is a crucial issue as we provide no information to the models on our aim to reconstruct value configurations from embeddings, by using only the generic prompt from the survey instructions (see Subsection \ref{subsec:prompt}) or by using no prompt at all. Furthermore, our best performing model "Linq-Embed-Mistral" has only moderate size with 7.11B parameters and performs well in the diverse tasks of the "Massive Text Embedding Benchmark" \cite{muennighoff_mteb_2023}, which points to its general capabilities.

The notion of "similarity" is actually not well-defined as a classical discussion in philosophy shows. The argument goes against the existence of a strong logical underpinning of notions of similarity, while its practical relevance is acknowledged. Similarity is considered "still serviceable in the streets" \cite{Goodman1972-GOOSSO}. Some parts of the original logical argument have been debated and countered \cite{leitgebNewAnalysisQuasianalysis2007}. We further want to note that our metaphor of semantic embeddings as human rater judgments helps us to prevent some concerns: Each dimension of semantic embeddings as the equivalent to a human rater judgment provides the necessary context ("similar to what?") to justify relative assessments of similarity for our use case.

\section*{Ethics Statement}

% Benefits: If the technology functions as intended, who benefits?

The technology of using semantic embeddings in research in ways that are similar to human rater judgments can become a common and established practice (as proposed for example by \citealp{guenoleEnhancingScaleDevelopment2025}). For researchers, we see obvious benefits as they will be provided with a novel way of gathering and analyzing data. This approach potentially promises to capture a richer tapestry of human behavior and experience than conventional methods relying on WEIRD (Western, Educated, Industrialized, Rich, Democratic) college student populations. In addition to such convenience samples, human rater judgments are typically sourced from crowdworkers that carry out tasks at often low hourly wages. There is reason to hope that crowdworking platforms react to technology changes with a quality initiative that leads to better opportunities for crowdworkers to earn adequate incomes.

% Harms: If the technology functions as intended,
% or if it fails, who might be harmed, and how? As always, results have to double and triple-checked for correctness and robustness.

Still, crowdworking delivers income to people around the world that come from many marginalized groups. Although it can leave workers underpaid, it is an integral and existential part of many lives. Failures of platforms to react and dwindling numbers of crowdworking jobs (that are instead outsourced to LLMs) can leave many people worse off than before.

On another, more distant future, level, we could also see some danger in people accepting analyses based on semantical embeddings as some kind of "gold standard", a LLM normality to compare with the results of humans, forgetting about many possible biases. Abstractly, like any assessment, this could increase the danger of diagnosing somebody as abnormal vis-a-vis an unattainable standard. 

% Vulnerabilities: Are any of the possible harms
% you've identified likely to fall disproportionately on populations that already experience marginalization or are otherwise vulnerable?

Generally, we see potential harm to fall mostly proportional on all parts of the population in terms of race, gender or class. However, potential lack of jobs for crowdworkers can be expected to have a disproportionate effect on people on low incomes in less affluent (and predominantly English-speaking) parts of the world. In addition, we try to limit the potential harm of creating novel synthetic LLM norms by restricting ourselves to a questionnaire of value orientations such as the PVQ-RR that is rather uncontroversial and is not positing any "correct" or "normal" value orientations for individuals.

% \section*{Acknowledgments}

% We provide a code and data \href{ADD ANO REPO}{repository} containing documented scripts that allow to fully repeat our analyses using openly available data. All human data that we analyze is retrieved from the \href{https://osf.io/w9as3/?view_only=e1f02bf232c34d39b9884398b4f2df63}{open paper repository} of \citet{schwartzMeasuringRefinedTheory2022}. We thank Shalom H. Schwartz and Jan Cieciuch for providing excellent replication materials. Figure \ref{fig:workflow} has been designed using resources from \href{https://www.flaticon.com/}{Flaticon.com}. We used Claude 3.7 Sonnet to help us write parts of the abstract for this work.

\bibliography{custom}

\appendix

\begin{figure*}[ht!]
\centering
  \includegraphics[width=0.6\linewidth]{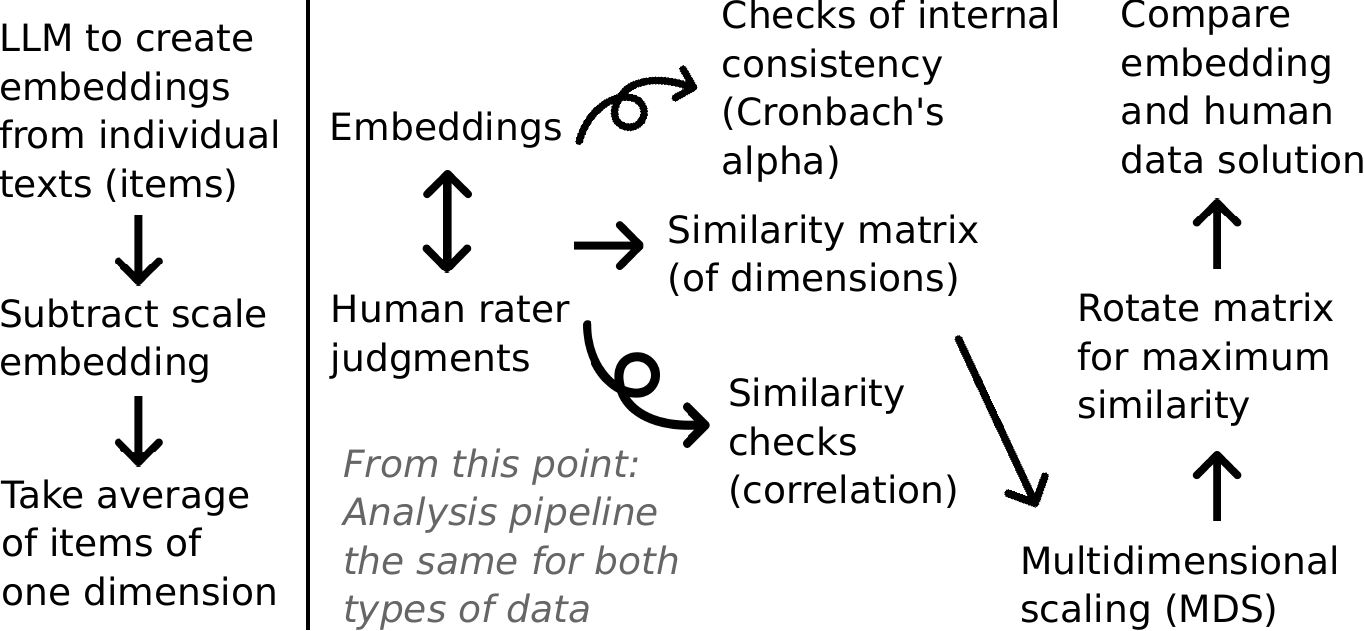}
  \caption{\label{fig:workflow} \textbf{Schematic overview of our workflow.} We use LLMs to create embeddings, subtract an average embedding of all items from each item embedding and aggregate by dimension through averaging. For both the embeddings as well as the human data, we compute Cronbach's alpha and create similarity matrices to compare both types of data. We run multidimensional scaling, rotate the configurations to maximum similarity and compare the results both visually as well as quantitatively.}
\end{figure*}

\section{Appendix}
\label{sec:appendix}

In our paper repository\footnote{\href{https://github.com/maxpel/embeddings\_values}{https://github.com/maxpel/embeddings\_values}}, we share documented code that allows other researchers to fully replicate our results using openly available data.

\subsection{Older lines of research on survey item similarity}

For completeness sake, we want to mention that work on semantic similarity of test items goes back before the recent developments in NLP research of 2017/18, e.g. \citet{arnulfPredictingSurveyResponses2014}. The usual approach back then was to create item embeddings and to compare them with each other for similarity using latent semantic analysis (LSA) \cite{rosenbuschSemanticScaleNetwork2020,arnulfSemanticOntologicalStructures2021}. Schematically, researchers used singular value decomposition on the co-occurence matrix of terms within items. The creation of semantic networks of items is a similar, related approach \cite{christensenSemanticNetworkAnalysis2021}. Classic use cases for those approaches involve researchers to come up with sample items that are then used to find already existing scales containing similar items or to check for item redundancy.

\subsection{Different, but related uses of semantic embeddings as human rater judgments}

Researchers have used embeddings that they trained to predict rating scales as a way to analyze descriptive wordresponses by human participants \cite{kjellNaturalLanguageAnalyzed2022}. They find that a BERT model works well for this task, especially with word embeddings. With a slightly different twist, researchers used item embeddings together with human rater judgments of personality items to predict ratings of completely new items that had not been rated by any participant yet \cite{abdurahmanDeepLearningApproach2024}.

\subsection{\label{subsec:cronbach}Cronbach's Alpha of human rater judgments and embeddings}

We compute Cronbach’s Alpha according to the usual formula
$$\alpha = {\frac{k}{k-1} } \left(1 - {\frac{\sum_{i=1}^k \sigma^2_{y_i}}{\sigma^2_y}} \right)$$
where $k$ represents the number of items in the dimension, for the PVQ-RR $k=3$

$\sigma_{y_i}^2$ the variance associated with each item $i$

$\sigma_y^2$ the variance associated with the total scores $y = \sum_{i=1}^k y_i$\\
\newline
For the human rater judgments:
$$\mathbf{y}_i = (r_{i1}, r_{i2}, \ldots, r_{in})$$
where each $r_{ij}$ represents the rating given by human rater $j$ for item $i$.\\
\newline
For the embeddings:
$$\mathbf{y}_i = (e_{i1}, e_{i2}, \ldots, e_{in})$$
where each $e_{ij}$  represents the element $j$ of the embedding vector for item $i$.

\subsection{\label{subsec:pca}PCA instead of MDS}

\begin{figure}[htpb!]
\centering
  \includegraphics[width=0.65\columnwidth]{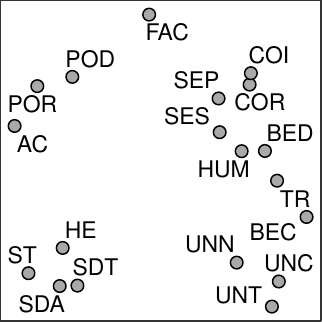}
  \caption{\label{fig:pca} \textbf{Replacing MDS with PCA.} We plot the first two dimensions found through PCA with the embeddings. We see similar clustering as compared to the MDS configuration with less overall circular structure.}
\end{figure}

As a robustness check, we replace MDS with PCA\footnote{\textbf{prcomp} in R with 'center = TRUE' and 'scale = TRUE'}. In Figure \ref{fig:pca}, we plot the first two dimensions that show similar clustering to the MDS solution but less circular structure since PCA does not guarantee to preserve distances.

\begin{figure}[htpb!]
\centering
  \includegraphics[width=0.65\columnwidth]{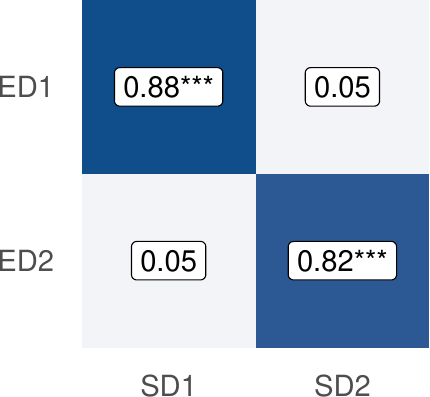}
  \caption{\label{fig:congruence} \textbf{Congruence coefficients used to measure the similarity between our two MDS configurations.} Labels refer to \textbf{E}mbeddings from linq and human rater judgments from \textbf{S}chwartz \cite{schwartzMeasuringRefinedTheory2022}.}
\end{figure}

\begin{figure*}[t!]
  \includegraphics[width=0.48\linewidth]{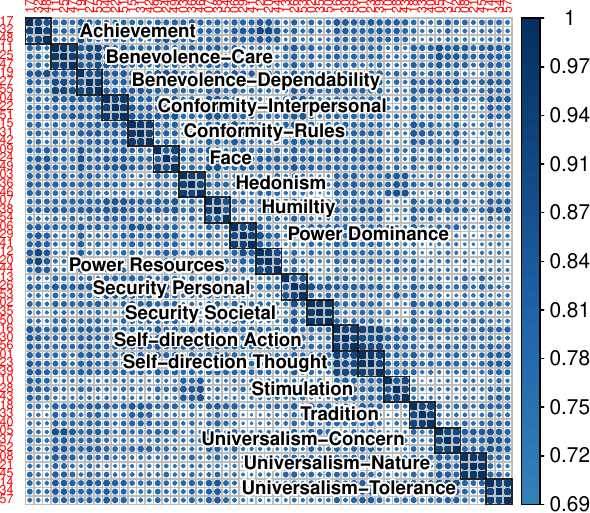} \hfill
  \includegraphics[width=0.48\linewidth]{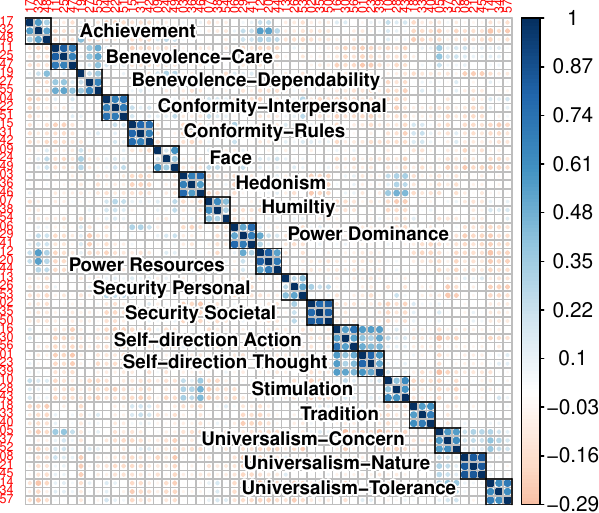}
  \caption{\label{fig:scalesubtraction} \textbf{Effect of questionnaire embedding subtraction on the item similarity matrix.} Left hand figure shows the raw embeddings, right hand figure the SQuID treated embeddings. By using SQuID to normalize for common standard features of natural language use, negative correlations appear.}
\end{figure*}

\subsection{\label{subsec:congruence}Congruence coefficients}

Congruence coefficients are computed for each of the two dimensions after Procrustes transformation of the embeddings configuration. There is no strict standard, but according to typical interpretations, values in the range of $0.85 \text{-} 0.94$ between two configurations indicate fair similarity \cite{lorenzo-sevaTuckersCongruenceCoefficient2006}. To test significance, we create pairs of 1000 random vectors, use Procrustes transformation to maximize their similarity and compute cosine similarity. We repeat this procedure 1000 times and find $p<0.001$ for the cosine similarities we report over all runs, indicating that they are significantly different from random baselines and cannot be explained as artifacts of Procrustes transformation.

Table \ref{tab:congruence_models} presents these scores for all five models tested.
For matching dimensions (SD1\_ED1 and SD2\_ED2), higher congruence coefficients indicate better alignment with the human-derived structure. For non-matching dimensions (SD1\_ED2 and SD2\_ED1), lower values indicate proper orthogonality. Consistent with our other evaluation measures in the main text such as Cronbach's Alpha, linq achieves the strongest performance with congruence scores of 0.88 and 0.82 for the two matching dimensions, substantially outperforming other approaches. Gemini also performs quite well, while the other models show weaker congruence.

\begin{table}[!htbp]
\fontsize{12.0pt}{14.4pt}\selectfont
\begin{tabular*}{\linewidth}{@{\extracolsep{\fill}}l|rrrr}
\toprule
 & \makecell{SD1\\ED1} & \makecell{SD1\\ED2} & \makecell{SD2\\ED1} & \makecell{SD2\\ED2} \\ 
\midrule\addlinespace[2.5pt]
gemini & 0.72 & \textbf{0.05} & \textbf{0.05} & 0.78 \\ 
jina & 0.37 & 0.24 & 0.26 & 0.78 \\ 
kalm & 0.61 & 0.13 & 0.14 & 0.57 \\ 
linq & \textbf{0.88} & \textbf{0.05} & \textbf{0.05} & \textbf{0.82} \\ 
mpnet & 0.53 & 0.29 & 0.31 & 0.26 \\ 
\bottomrule
\end{tabular*}
\caption{\label{tab:congruence_models}\textbf{Factor congruence scores with human rater judgments for all models.} For matching dimensions (SD1\_ED1 and SD2\_ED2) higher is better. For differing dimensions (SD1\_ED2 and SD2\_ED1) lower is better.}
\end{table}

\subsection{\label{subsec:corrplot_dimensions}Effect of Questionnaire Embedding Subtraction}

\begin{figure}[!htpb]
  \includegraphics[width=\columnwidth]{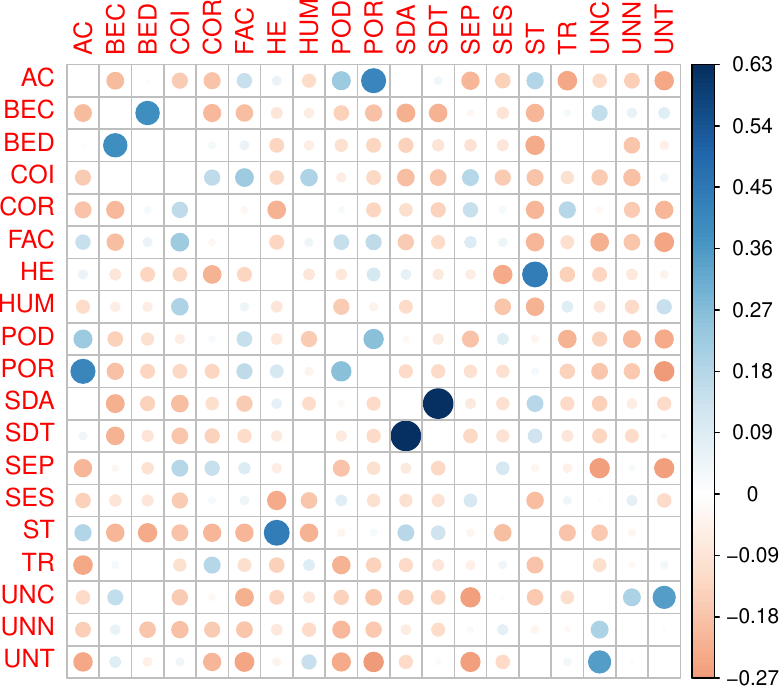}
    \caption{\label{fig:correlationplots} \textbf{Similarity matrix of SQuID treated embeddings averaged by dimension.} We compute correlation coefficients between SQuID treated embeddings that are averaged by dimension. We don't show the diagonal of 1s to better allow to observe correlation ranges.}
\end{figure}

Figure \ref{fig:scalesubtraction}, from left to right, shows visually that with SQuID we achieve negative correlations in the item-item matrix (dimensions sorted alphabetically).

\subsection{\label{subsec:prompt}Prompt}

For all models that allow to include a prompt, we use the following wording taken exactly as-is from the specification of the \href{https://osf.io/w9as3/files/osfstorage/5f436f92f579150161ea889c}{English version} of the PVQ-RR: "Here we briefly describe different people. Please read each description and think about how much that person is or is not like you. Put an X in the box to the right that shows how much the person described is like you."

\end{document}